\documentclass[12pt]{l4dc2021}

\usepackage{times}
\usepackage{booktabs}
\usepackage{adjustbox}
\usepackage{algorithm}
\usepackage{algpseudocode}

\usepackage{balance}
\usepackage{dirtytalk}
\usepackage{amsmath}

\newcommand\numberthis{\addtocounter{equation}{1}\tag{\theequation}}
\DeclareMathOperator{\sign}{sign}

\title[Spatio-Temporal Specifications: Learning and Synthesis]{ Learning Spatio-Temporal Specifications for Dynamical Systems}

\author{%
 \Name{Suhail Alsalehi}$^{1}$ \Email{alsalehi@bu.edu}
 \AND
 \Name{Erfan Aasi}$^{2}$ \Email{eaasi@bu.edu}
  \AND
   \Name{Ron Weiss}$^{3}$ \Email{rweiss@mit.edu}
  \AND
  \Name{Calin Belta}$^{1,2}$ \Email{cbelta@bu.edu}\\
 \addr ${1}$: Division of Systems Engineering, Boston University, Boston, MA, USA \\
 \addr $2$: Mechanical Engineering Department, Boston University, Boston, MA, USA\\
  \addr $3$: Biological Engineering Department, Massachusetts Institute of
Technology, Cambridge, MA, USA
}
\begin{document}
\maketitle

\vspace{-10mm}
\begin{abstract}
Learning dynamical systems properties from data provides important insights that help us understand such systems and mitigate undesired outcomes. In this work, we propose a framework for learning spatio-temporal (ST) properties as formal logic specifications from data. We introduce SVM-STL, an extension of Signal Signal Temporal Logic (STL), capable of specifying spatial and temporal properties of a wide range of dynamical systems that exhibit time-varying spatial patterns. Our framework utilizes machine learning techniques to learn SVM-STL specifications from system executions given by sequences of spatial patterns. We present methods to deal with both labeled and unlabeled data. In addition, given system requirements in the form of SVM-STL specifications, we provide an approach for parameter synthesis to find parameters that maximize the satisfaction of such specifications. Our learning framework and parameter synthesis approach are showcased in an example of a reaction-diffusion system.  
\end{abstract}


\section{Introduction and Related Works}

Many dynamical systems exhibit time-varying spatial behaviors. Smart cities, robotics swarms, and biological multi-cellular systems are just a few examples. With the increasing complexity of such systems, there is a need for formal ways to describe their spatial and temporal properties. To be deemed useful, these properties must be interpretable for humans and amenable to rigorous mathematical analysis. Two of the main challenges are inferring such properties from data (\textit{the inference problem}) and synthesizing system parameters (inputs) such that certain properties are met (\textit{the parameter synthesis problem}). This work explores both problems for dynamical systems.

Machine learning and formal logics are two fields with research efforts on the inference problem. Deep Neural Networks (DNN) have shown success in inferring properties from data (feature extraction). However, the inferred properties lack interpretability and can only be used by machine learning models for tasks such as classification. On the other hand, due to the expressivity and readability,  formal logics are widely used for specifying spatial and temporal properties of dynamic systems. Inferring formal logic specifications from system executions has been explored in the literature, e.g.~\cite{asarin2011parametric, hoxha2018mining, bombara2016decision, vazquez2017logical, jha2019telex,swarmstl,Fan2020,mining_STREL,gtl,wgtlnn,swarmstl,STSL}. 

Spatio-temporal (ST) logics are formal languages capable of specifying ST properties of dynamical systems~\cite{spatel,SaSTL,STSL,STREL,STREL_suhail,swarmstl}. A common theme among these works is combining spatial logics with temporal logics to produce ST logics. For example, the authors of~\cite{Mehdipour2018} nest spatial properties in STL formulae by defining predicates as geometric distances to hyperplanes of SVM classifiers, providing a qualitative valuation for spatial patterns. 
The literature on ST logics focuses on time-varying spatial patterns given by graphs, e.g.~\cite{STREL}, quadtrees~\cite{spatel}, abstractions of systems,  e.g.~\cite{STSL} and so on. However, there are no works on specifications for time-varying spatial patterns given by images.

To address this concern, we introduce SVM-STL, a logic capable of describing ST properties of dynamical systems. SVM-STL nest spatial properties of images into STL by defining machine learning-based predicates that automate feature extraction from \textit{ST trajectories} (sequences of spatial patterns generated by executions of a system). SVM-STL is equipped with qualitative semantics that describes whether a trajectory satisfies a SVM-STL formula or not, as well as quantitative semantics, which quantifies the degree of satisfaction of a formula by a ST trajectory.

We provide a framework for learning SVM-STL formulae from ST trajectories by separating the learning of spatial properties from the learning of temporal properties. First, we ignore the temporal aspect of the data and utilize machine learning techniques to learn predicates which capture the spatial properties in the data. The predicates are a combination of SVM binary classifiers and a neural network model for automated feature extraction from images. Then, we utilize a decision tree-based algorithm~\cite{aasi2021classification} to learn SVM-STL formulae from ST trajectories. We also provide an unsupervised learning approach to learn SVM-STL formulae from unlabeled ST trajectories. To the best of our knowledge, this is the first framework for learning formal logics specifications from ST trajectories, where spatial patterns are given by images.

Synthesizing parameters from spatio-temporal specifications has been the focus of several works in the literature, e.g.~\cite{ Haghighi2019-as, Liu2018, bozkurt2020control,STREL_suhail}. In this work, we introduce our approach to parameter synthesis for systems with spatial and temporal requirements. Our approach is unique in that we can learn requirements from executions with desired behavior. The efficacy of the learning framework and the parameter synthesis solution are showcased in a case study of a reaction-diffusion system. 

\section{Preliminaries and Notations}
\label{sec:prelim}

Let $\mathbb{R},\mathbb{Z},\mathbb{Z}_{\geq 0}$ be the set of real numbers, integers, and non-negative integers, respectively. 
Given $a, b \in \mathbb{Z}_{\geq 0}$, with slight abuse of notation, we write $[a, b] = \{k \in \mathbb{Z}_{\geq 0} | a \leq k \leq b\}$.
A \textbf{signal} $s$ is a function $s: \mathbb{T} \rightarrow \mathbb{R}^{n}$ that maps each discrete time point $k \in \mathbb{T} = [0, T]$, $T \in \mathbb{Z}_{\geq 0}$, to a n-dimensional real-valued vector $s[k] \in \mathbb{R}^n, n \in \mathbb{Z}_{\geq 0}$.
A RGB image (\textbf{image}) is a phenotypical observation of a system at a fixed time point. RGB is an additive color model~\cite{image_def} in which the channels red, green, and blue are added together in various ways to reproduce a broad range of colors. An image is denoted by $I\in \mathbb{R}^{L\times W \times C}$, where $L,W,C \in \mathbb{Z}_{\geq 0}$ are the length, width and channels. 
A \textbf{spatio-temporal (ST) trajectory} $S$ is a function $S: \mathbb{T} \rightarrow \mathbb{R}^{L\times W \times C}$ that maps each discrete time point $k \in \mathbb{T}$, to an image $S[k]\in \mathbb{R}^{L\times W \times C}$, where $L,W,C \in \mathbb{Z}_{\geq 0}$.

\subsection{Convolutional Neural Network (CNN)}
\label{sec:cnn}

CNN is a deep learning algorithm commonly used for analyzing images. A simple CNN consists of one or more convolutional and fully-connected layers. Convolutional layers extract the high-level features from images, while fully connected layers learn classifiers, regression models, etc. Transfer learning~\cite{transfer_learning}, i.e. storing knowledge gained while solving one problem and applying it to a different but related problem, is often done to save time, computational power, or when there are not enough examples to train a new model. For instance, convolutional layers of a CNN that is trained on a broad set of classes (e.g., AlexNet~\cite{alexnet}, VGG16~\cite{vgg16} and Inception~\cite{inception}) are good feature extractors~\cite{VGG16_SVM2,VGG16_SVM1,cnnAsExtractors,rs10071119}. In this work, we use a pre-trained CNN as a \textbf{feature extractor}, which is given by a function $f_{cnn}: \mathbb{R}^{L\times W\times C}\rightarrow \mathbb{R}^m$. 

\section{SVM-STL Specifications}
\label{sec:svm-stl}

We introduce SVM-STL, a formal language capable of specifying spatial and temporal properties of dynamical systems such as \say{(\textit{eventually} in the time interval [0,30] large spots are observed for 10 time steps) AND (\textit{Never} in time interval [0,30] small spots are observed)}(see another example in Fig.~\ref{fig:par_synth}). SVM-STL takes inspiration from~\cite{Mehdipour2018} and builds upon Signal Temporal Logic (STL)~\cite{original_stl}. The syntax of SVM-STL formulae is defined over $S$ as 
\begin{equation}
\label{eq:svm-stl-syntax} \nonumber
\varphi := \top | \mu_j | \neg \varphi | \varphi_1 \wedge \varphi_2 | \varphi_1 \vee \varphi_2 | G_{[a,b]} | F_{[a,b]}
\end{equation}
where $\varphi,\varphi_1,\varphi_2$ are STL formulae, $\top$ is Boolean \textit{True}; $\mu_j$ is a \textit{predicate} of the form $\mu_{j}:= h_j(S[k]) \sim r $, defined over components of ST trajectories $S$, where $h_j:\mathbb{R}^{L\times W \times C} \rightarrow  \mathbb{R}, j = 1,\ldots,n$ is a \textit{predicate function}, $\sim \in\{> ,\leq \}$ and $r\in \mathbb{R}$ is a threshold; 
$\neg,\wedge$, and $\vee$ are the logical operators \textit{negation} , \textit{conjunction}, and \textit{disjunction}, respectively;
and $F_{[a,b]}, G_{[a,b]}$ are the temporal operators \textit{eventually}  and \textit{always}, with $a,b \in \mathbb{Z}_{\geq 0}$.

A predicate function $h_j$ is a classifier that maps images to real-values corresponding to how strong images belong to (spatial) class $j$. The mathematical structure of $h_j$ is introduced in Sec.~\ref{sec:learn_predicate}. 

SVM-STL is equipped with qualitative and quantitative semantics. The qualitative semantics describes satisfaction/violation of a SVM-STL formula $\varphi$ at time $k$ by a ST trajectory $S$ and we use $S[k] \models \varphi$ to denote (Boolean) satisfaction. The qualitative semantics is given recursively by: 
\begin{align*}
(S,k)&\models\mu_j & \Leftrightarrow\ & h_j(S[k]) \sim r \\
(S,k)&\models\neg\varphi & \Leftrightarrow\ & \neg((S,k)\models\varphi) \\
(S,k)&\models\varphi_{1}\wedge \varphi_{2} & \Leftrightarrow\ & (S,k)\models\varphi_{1}\wedge(S,k) \models \varphi_{2} \\
(S,k)&\models{F}_{[a, b]} \varphi & \Leftrightarrow\ & \exists k' \in[k+a, k+b] \text { s.t. }(S,k')\models\varphi \\
(S,k)&\models{G}_{[a, b]} \varphi & \Leftrightarrow\ & \forall k' \in[k+a, k+b],(S,k')\models\varphi \numberthis \label{eq:qual_semantics}
\end{align*}


Similar to~\cite{donze2010robust}, the quantitative semantics is given by the real-valued robustness function $\rho(S, \varphi,k)$, which captures the degree of satisfaction of a formula $\varphi$ by a ST trajectory $S$. Specifically, a positive robustness score ($\rho(S, \varphi,k) \geq 0$) implies satisfaction $S \models\varphi$, while negative robustness ($\rho(S, \varphi,k) < 0$) implies violation. Given a formula $\varphi$ and a ST trajectory $S$, the robustness at time $k$ is recursively defined as follows: 
\begin{align*}
\label{eq:stl-org}
        \rho (S,\mu_j,k)  & = \begin{cases}
    		            	h_j(S[k])-r, & \text{if $\mu := h_j(S[k]) > r$}\\
                            r-h_j(S[k]), & \text{otherwise}
    		                \end{cases}\\
        \rho (S,\neg \varphi,k)  & = - \rho (S,\varphi,k)\\
        \rho (S,\varphi_1 \vee \varphi_2,k) & =  \max (\rho (S,\varphi_1,k), \rho (s,\varphi_2,k))\\
        \rho (S,F_{[a,b]} \varphi,k) & =  \max_{k'\in [k+a,k+b]} \rho(S,\varphi,k')\\
        \rho (S,G_{[a,b]} \varphi,k) & = \min_{k'\in [k+a,k+b]} \rho  (S,\varphi,k')
\end{align*}

Similar to~\cite{asarin2011parametric}, we define \textbf{Parametric SVM-STL (PSVM-STL)}, which is an extension of SVM-STL, where the time bounds $a, b$ of temporal operators and threshold $r$ of the predicate are parameters. The set of all possible valuations of all parameters in a PSVM-STL formula $\varphi$ is called the parameter space and is denoted by $\Theta$. A particular valuation of a PSVM-STL formula $\varphi$ at $\theta \in \Theta$ is denoted by $\varphi_{\theta}$.

\textbf{PSVM-STL primitives} are simple PSVM-STL formulae. We define the set of first-order primitives as $\mathcal{P} = \{F_{[a,b]}(h_j(S[k]) \sim r), G_{[a,b]}(h_j(S[k]) \sim r)\}$, where $r \in \mathbb{R}$; $a,b,c \in \mathbb{Z}_{\geq 0}$; and $\sim \in \{\leq , > \}$. The parameters of $\mathcal{P}$ are $(r,a,b)$.

\textbf{Weighted SVM-STL} (wSVM-STL) is another extension of SVM-STL (based on~\cite{wSTL}) with the same qualitative semantics as SVM-STL, but its robustness degree is modulated by the weights associated with the Boolean and temporal operators. In this paper, we focus on a fragment of wSTL, with weights on conjunctions only, i.e. $\bigwedge_{i=1,..,N}^{\mathbf{w}} \varphi_i$ and $\bigvee_{i=1,..,N}^{\mathbf{w}} \varphi_i$. The weight ${\mathbf{w}}=[w_1,..,w_N]$ assigns a positive weight to each subformula $\varphi_i$. The weights $\mathbf{w}$ capture the importance of specifications for conjunctions or priorities of alternatives for disjunctions.

\section{Learning Spatio-temporal Properties}
\label{sec:learning-main}

\textbf{The Inference Problem:} \textit{Given a set  $\mathbf{S} = \{S^{(i)}\}_{i=1}^{N_S}$ representing executions of a system (ST trajectories), learn ST properties of the system in the form of SVM-STL formulae.}

We provide a framework to infer SVM-STL formulae from system executions in two stages. In the first stage, we construct predicate functions $h_1,..,h_{n_I}$ that capture the spatial properties in the data. In the second stage, we learn SVM-STL formulae using a decision tree-based approach to capture the ST properties of the data. To accommodate learning from unlabeled images and trajectories, we utilize unsupervised learning techniques to cluster and label images and trajectories. Overall, we divide the learning problem into four sub-problems that we address in the subsequent sections: 1)
 clustering spatial data (images), 2) learning spatial properties, 3) clustering ST trajectories, 4) learning spatio-temporal specifications.

\subsection{Clustering Images}
\label{sec:cluster_images}

Given a set of data points, one can use unsupervised clustering algorithms to organize unlabeled data into $n$ similarity groups called clusters. Clustering requires a distance/similarity measure $d$, a criterion function $f_{crit}$ and an algorithm to optimize the criterion function. The choice of $n,d,f_{crit}$ and the clustering algorithm depends on the type of data and purpose of clustering. In this work, we use the k-means clustering algorithm which aims to partition data points into $n$ clusters in which each data point belongs to the nearest mean $\mu_i$ (cluster center) while minimizing the criterion function. 

Let $\mathbf{I} = \{I^{(i)}\}_{i=1}^{N_{I}}$ be a set of images. We consider the feature extractor $f_{cnn}:\mathbb{R}^{W \times L \times C} \rightarrow \mathbb{R}$ and criterion function $f_{crit}$ given by
$f_{crit}(I,\mu) = \sum_{\mu_j \in \mu} \sum_{I \in c_j} d_I(I,\mu_j)$, where $c_j = \{I^{(i)}| j=arg\min_{j=1,..,n_I}(d_I(I^{(i)},\mu_j)) \}$; $\mu =\{\mu_1,..,\mu_{n_I}\}, \mu_1,..,\mu_{n_I} \in \mathbb{R}^m$; and $d_I$ is the similarity measure given by $d_I(I,\mu_j) = || f_{cnn}(I) - \mu_j||^2$. We want to find the best set of cluster centers $\mu^{best} $ that minimize the objective function $f_{crit}$.  
\begin{equation}
    \label{eq:cluster_images}
    \mu^{best} =  arg \min_{\mu} (f_{crit} (\mathbf{I},\mu));
\end{equation}

To find $\mu^{best}$, we use PSO~\cite{kennedy1995particle} which optimizes the problem by iterative improvement of a candidate solution according to a criterion function. The PSO-based solution is summarized in Alg.~\ref{alg:cluster_images}, which starts by randomly initializing a set of $K$ particles with positions (parameters) $\pi_k = \{\pi_{k,j}\}$, where $\pi_{k,j} \in \mathbb{R}^{m}, k =1,..,K$, and $j =1,..,n_I$; and velocities $v_k = \{v_{k,j}\}$, where $v_{k,j} \in \mathbb{R}^{m}$. Each particle represents a candidate solution to~\eqref{eq:cluster_images}. At each iteration, the criterion function is evaluated for $K$ sets of cluster centroids $\pi_k, k =1,..,K$. The position of the $k$-th particle with the best set of centers so far is stored in the variable $\pi_k^{best}$. Similarly, the position that performed best (lowest $f_critic$) among all particles so far is stored in the variable $\pi^{best}$. At the end of each iteration, positions and velocities of particles are updated as follows: 
\begin{align}
\label{eq:pso1}
    v_k &\gets W v_k + \eta(0,r_p) (\pi_k^{best}-\pi_k) +\eta(0,r_g) (\pi^{best}-\pi_k) \\
\label{eq:pso2}
    \pi_k &\gets \pi_k +v_k
\end{align}
where $\eta(a,b)$ generates a random number from a uniform distribution in the interval $[a,b]$ and $W,r_p,r_g \in \mathbb{R}$ are the PSO hyperparameters (\cite{kennedy1995particle}) that are specified by the user. The algorithm keeps iterating  until a stopping condition $Stop$ is met, e.g. after a certain robustness threshold or if $\pi^{best}$ does not change significantly over the last $z$ iterations. Once the stopping condition is met, each image $S^{(i)}$ is given a label $l^{(i)}$. Finally, we construct the new set $\mathbf{\mathcal{I}} = \{(I^{(i)},l_I^{(i)})\}_{i=1}^{N_{I}}$, which consists of images $I^{(i)}$ and their labels $l_I^{(i)} \in \{1,..,n_I\}$. Note that domain knowledge is incorporated to determine the number of clusters $n_I$ and make decisions to merge or drop certain clusters. 

\begin{algorithm}[htb]
\caption{Clustering images using PSO}
\begin{algorithmic}[1]
    \State \textbf{Input:} $\mathbf{I} = \{I^{(i)}\}_{i=1}^{N_{I}},f_{cnn},d_I,n_I,f_{crit}$, (\textit{hyperparameters} : $W,r_p,r_g,K$)
    \State \textbf{Initialize:} $\pi_{k},\pi_k^{best},\pi^{best}, \mathbf{v}_{k}; k = 1,..K$ \Comment{initialize particles} \label{}
    \State \textbf{while} $\neg Stop$ \Comment{terminate if stopping condition is met}
    \State \hskip1.5em \textbf{for $k:= 1,..,m$ do}
    \State \hskip1.5em \hskip1.5em$ \pi_{k}^{best}\gets arg\min_{\pi = \{\pi_{k},\pi_k^{best}\}} f_{crit}(\mathbf{I},\pi) $ \Comment{best set of centers for particle $k$ so far}
    \State \hskip1.5em \hskip1.5em $ [\pi_k,v_k] \gets $ update according to \eqref{eq:pso1} and \eqref{eq:pso2}
    \State \hskip1.5em $ \pi^{best} \gets arg\min_{\pi = \{\pi^{best},\pi_1^{best},..,\pi_m^{best}\}} f_{crit}(\mathbf{I},\pi)  $  \Comment{best set of centers overall so far}
    \State $\mu^{best} \gets \pi^{best}$
    \State $l_I^{(i)} = arg\min_{j}(d_{I}(I^{(i)},\mu^{best}))$
    \State \textbf{Return: } $\mathbf{\mathcal{I}}  = \{(I^{(i)},l_I^{(i)})\}_{i=1}^{N_{I}}$
\end{algorithmic}
\label{alg:cluster_images}
\end{algorithm} 

\subsection{Learning Spatial Properties} 
\label{sec:learn_predicate}

In this section, we construct the predicate functions $h_1,..,h_{n_I}$ that capture the spatial properties in the set of labeled images  $\mathbf{\mathcal{I}}  = \{(I^{(i)},l_I^{(i)})\}_{i=1}^{N_{I}}$, where $l_I^{(i)} \in \{1,..,n_I\}$ is the label of image $I^{(i)}$. As stated in Sec.~\ref{sec:svm-stl}, a predicate function $h_j$ takes an image as an input and returns a real-value corresponding to how strongly the image belongs to the class $j$. We propose a modified version of Support Vector Machine (SVM) as predicate functions.

SVM is a widely used binary classification algorithm, known for its simplicity, high speed, and accuracy in multi-dimensional spaces. Since SVM classifiers do not support tasks with more than two classes, we split the multi-class classification set of samples into multiple binary classification sets (One-vs-Rest) and fit a binary classification model on each. Specifically, we divide the set $\mathbf{\mathcal{I}}$ into $n_I$ sets $\mathbf{\mathcal{I}}_{1},..,\mathbf{\mathcal{I}}_{{n_I}}$ where $\mathbf{\mathcal{I}}_{j} = \{(I^{(i)},l_{B})|l_{B} = 1, \text{ if } l_I^{(i)} = j \text{, and } -1 \text{ otherwise} \}$, where $l^{(i)}_B$ is the Boolean label. Thus, problem above is reduced to $n_I$ binary classification problems. 

Next, we consider a set $\mathcal{I}_j = \{(I^{(i)},l_{B}^{(i)})\}_{i=1}^{N_{I}}$ of $N_I$ training samples, where $I^{(i)}$ is the $i^{\text{th}}$ observation and $l_{B}^{(i)} \in \{-1, 1\}$ is its label. We want to  finds a decision boundary of the form $\omega_j^T f_{cnn}(I)+b_j = 0$ that maximizes the geometric margin between support vectors. Formally, the optimization problem can be written as (\cite{gunn1998support}):
\begin{equation}
    \label{eq:learn_svm}
     \min_{\omega_j}  ||\omega_j|| \, \text{, subject to } l_i(\omega_j^T f_{cnn}(I^{(i)})+b_j)\geq 1, \: i=1,..,N_I.
\end{equation}

This is a convex optimization problem that can be solved using a gradient-based method. The resulting SVM classifier is given by $y_j(I) = sign (\omega_j^T f_{cnn}(I) + b_j)$, with $y_j:\mathbb{R}^{W \times L \times C} \rightarrow \{-1, 1\}$. We define the predicate function as the signed Euclidean distance from images to the decision boundary of the learned classifiers. Specifically, the predicate function $h_j$ is given by 
\begin{equation}
\label{eq:svm_dist}
    h_j(I^{(i)}) =  \frac{\omega_j^T f_{cnn}(I^{(i)}) +b_j }{||\omega_j||},
\end{equation}
where if $h_j(I)>0$ then  $l_{B}^{(i)} = 1$, otherwise $l_{B}^{(i)} = -1$. 

With slight abuse of notation, we define the operator
$h: S \rightarrow s$, which maps ST trajectories $S: \mathbb{T} \rightarrow \mathbb{R}^{L \times W \times C}$ to \textbf{spatio-temporal (ST) signals} $s: \mathbb{T} \rightarrow \mathbb{R}^{n_I}$, a signal-like representation of ST trajectories.

\subsection{Clustering ST trajectories}
\label{sec:cluster_ST}

Let $\mathbf{S} = \{S^{(i)}\}_{i=1}^{N_S}$ be a set of ST trajectories and consider the operator $h$. We consider the criterion function $f_{crit}$ given by
$f_{crit}(S,\mu) = \sum_{\mu_j \in \mu} \sum_{S \in c_j} d_{dtw}(h(S),\mu_j)$, where $c_j = \{S^{(i)}| j=arg\min_{j=1,..,n_S}(d_{dtw}(S^{(i)},\mu_j)) \}$ and $d_{dtw}: s \times s \rightarrow R$ is the dynamic time wrapping distance - a similarity measure between two one-dimensional temporal sequences~\cite{dtw}. For $n$ dimensional sequences, we rearrange sequences into long one-dimensional sequences.

We follow a similar approach to clustering images (Sec.~\ref{sec:cluster_images}) to find the best set of cluster centers $\mu^{best} = \{\mu_1,..,\mu_{n_I}\}$, where $\mu_1,..,\mu_{n_I} \in \mathbb{R}^{m \times T} $ that minimize the objective function $f_{crit}$: 
\begin{equation}
    \label{eq:cluster_ST}
    \mu^{best} =  arg \min_{\mu} (f_{crit} (S,\mu)).
\end{equation}
We employ PSO to find the best set of cluster centers that minimizes the objective function $f_{crit}$. With appropriate changes, we use Alg.\ref{alg:cluster_images} to construct a new set $\mathbf{\mathcal{S}'} = \{(S^{(i)},l_S^{(i)})\}_{i=1}^{N_S}$ that consists of ST trajectories $S^{(i)}$ and their corresponding labels $l_S^{(i)} \in \{1,..,n_S\}$. A label signifies that a trajectory belongs to the class of that label. 

\subsection{Learning Spatio-temporal Specifications}
\label{sec:learn_SVM_STL}

Now that we have labeled ST trajectories, we want to learn ST properties in the form of SVM-STL formulae. For simplicity, we will consider the case of two classes (positive and negative).
Let $C = \{1, -1\}$ be the set of positive and negative classes. We consider the set  $\mathcal{S}' = \{(S^{(i)},l_S^{(i)})\}_{i=1}^{N_S}$, where $S^{(i)}$ is the $i^{\text{th}}$ trajectory and $l_S^{(i)} \in C$ is its corresponding label. Consider also the set of SVM-STL primitives $\mathcal{P}$, the predicate functions $h_1,..,h_{n_I}$ and the operator $h:S \rightarrow s$. We map ST trajectories in $\mathcal{S}'$ into ST signals using $h$ to produce the new set $\mathcal{S} = \{(s^{(i)},l_S^{(i)})\}_{i=1}^{N_S}$. We want to learn SVM-STL specifications $\varphi$ such that the misclassification rate $MCR(\varphi)$ is minimized. The misclassification rate is given by:
\begin{equation}
    MCR(\varphi) := \frac{|\{ s^{(i)}| (s^{(i)} \models \varphi \wedge l_S^{(i)} = -1) \vee (s^{(i)} \not\models \varphi \wedge l_S^{(i)} = 1) \}| }{N_S} ; i =1,..,N_s
\end{equation}

Inspired by the AdaBoost method~\cite{shalev2014understanding} and the boosted concise decision tree method in~\cite{aasi2021classification}, we use a Boosted Decision Tree (BDT) method to learn the SVM-STL formulas, explained in Alg.~\ref{alg:boosted_trees}. We do not consider any conciseness technique in the method, as the focus of our paper is not on the simplicity of the output formulas, and we mostly desire to obtain an acceptable MCR on the data by using the boosted nature of the method.

Here, we briefly explain the algorithm and its full description can be found in~\cite{aasi2021classification}. Alg.~\ref{alg:boosted_trees} takes as input the labeled dataset $\mathcal{S}$, the number of decision trees to grow $K$, and the shallow decision tree method $\mathcal{E}$ as the weak learning algorithm. An uniform distribution is assigned to the signals as an initial weight distribution (line~\ref{lst:line:initialize}). The algorithm iterates over the number of trees (line~\ref{lst:line:loop}) and at each iteration, the weak learning method $\mathcal{E}$ constructs a shallow decision tree $f_{DT}^k(.)$. The decision trees are constructed using first-order primitives $\mathcal{P}$ and the misclassification gain impurity measure~\cite{breiman1984classification}. The rest of the algorithm (lines~\ref{lst:line:misclasserror}-~\ref{lst:line:weightupdate}) follows the standard AdaBoost method, and the final classifier is constructed as the weighted sum of the shallow decision trees (line~\ref{lst:line:output}). 


\begin{algorithm}[htb]
\caption{Boosted Decision Trees (BDT)}
\begin{algorithmic}[1]
    \State \textbf{Input:} dataset $\mathcal{S} = \{(s^{(i)}, l^{(i)})\}_{i = 1}^{N}$, number of decision trees $K$, weak learning method $\mathcal{E}$
    \State \textbf{Initialize:} $\forall \, (s^{(i)},l^{(i)}) \in \mathcal{S}: \, D_1(s^{(i)}) = 1/N_S$ \label{lst:line:initialize}
    \State for k = 1, $\ldots$, K: \label{lst:line:loop}
    \State \hskip1.5em $\mathcal{E}(\mathcal{S}, D_k) \Rightarrow \text{classifier} \, f_{DT}^{k}(\cdot)$  \label{lst:line:sinlgtree}
    \vspace{1mm}
    \State \hskip1.5em $\epsilon_k \gets \sum_{(s^{(i)}, l^{(i)}) \in \mathcal{S}} D_k(s^{(i)}) \, \cdot \, \mathbf{1}[l^{(i)} \neq f_{DT}^{k}(s^{(i)})]$ \label{lst:line:misclasserror}
    \vspace{1mm}
    \State \hskip1.5em $\alpha_k = \frac{1}{2} \ln{(\frac{1}{\epsilon_k} - 1)}$ \label{lst:line:treeweight}
    \vspace{1mm}
    \State \hskip1.5em $D_{k+1}(s^{(i)}) \propto D_k(s^{(i)}) \exp{(-\alpha_k \cdot l^{(i)} \cdot f_{DT}^{k}(s^{(i)}))}$ \label{lst:line:weightupdate}
    \vspace{1mm}
    \State $f_{BDT}^{K}(\cdot) = \sign(\sum_{k=1}^{K} \alpha_k \cdot f_{DT}^{k}(\cdot))$ \label{lst:line:output}
    \vspace{1mm}
    \State \textbf{Return: } $f_{BDT}^{K}(\cdot)$ \Comment{final classifier}
\end{algorithmic}
\label{alg:boosted_trees}
\end{algorithm} 


\begin{remark}[Applicability to other logics]
The framework presented above can be generalized to learn formulae of any ST logic that is made of STL over spatial classifiers, whenever there is a way to learn correct (a spatial classifier is correct if positive values indicate image belong to the spatial class, and negative values indicate image does not belong to the spatial class) and interpretable spatial classifiers. Specifically, the method will work for any predicate functions $h_j: \mathbb{D} \rightarrow \mathbb{R}, j=1,..,n_I$ over the spatial domain $\mathbb{D}$, where $n_I$ is the number of spatial classes in the data. For example, the Tree Spatial Superposition Logic (TSSL)~\cite{TSSL} is a spatial logic equipped with quantitative semantics (robustness) that is interpretable, real-valued, and correct. TSSL learns specifications from states of networked systems with spatial domain $\mathbb{D} = \mathbb{R}^{m\times m}$. Thus, one can learn $n_I$ (one-vs-rest) spatial classifiers (TSSL formulae), and use the robustness functions as predicate functions $h_j:=\rho_j:\mathbb{D} \rightarrow \mathbb{R} $ . In this case, our framework can be applied with minimal modifications to learn ST properties in the form of TSSL-STL formulae. 
\end{remark}

\section{Parameter Synthesis}
\label{sec:par_synth}

Assume that certain desired ST properties were captured while inferring SVM-STL specifications from trajectories. One might want to find the set of system parameters (inputs) such that the executions from the system satisfy the desired specifications. In the following, we provide our approach to parameter synthesis from SVM-STL specifications.  

Consider a system $\mathbf{S}$ that exhibits time-varying spatial patterns depending on $p$ design parameters $\pi \in \Pi\subset \mathbb{R}^{p}$ (see Sec. \ref{sec:experiments} for an example). 
The ST trajectory generated by parameters $\pi$ is denoted by $S_{\pi}$. Consider also some ST property given as a SVM-STL formula $\varphi$. We want to find parameters $\pi^*$ such that the specification given by $\varphi$ is maximally satisfied, i.e.
\begin{equation}
    \label{eq:parameter_synthesis}
    \pi^* =  arg \max_{\pi} (\rho (\varphi,S_{\pi},0));\pi \in \Pi
\end{equation}
Note that the objective function is, in general, not differentiable. Heuristic optimization algorithms such as genetic algorithms, particle swarm optimization (PSO), or simulated annealing can be used to solve the optimization problem. Our PSO-based solution to~\eqref{eq:parameter_synthesis} is summarized in Algorithm.~\ref{alg:par_synth}. PSO starts by randomly initializing a set of $K$ particles with positions (parameters) $\pi_k \in \Pi$ and velocities $v_k \in V \subset \mathbb{R}^{p}$, $k=1,..,K$. Each particle represents a candidate solution to~\eqref{eq:parameter_synthesis}. At each iteration, $K$ ST trajectories are generated and the robustness of each trajectory is evaluated. The position of the $i$th particle with the best performance so far is stored in the variable $\pi_k^{best}$. Similarly, the position that performed best (highest robustness) among all particles so far is stored in the variable $\pi^{best}$. At the end of the iteration, the position and velocity of each particle are updated according to \eqref{eq:pso1} and \eqref{eq:pso2}. The algorithm keeps iterative until a stopping condition $Stop$ is met.

\begin{algorithm}[ht!]
\caption{\footnotesize Parameter Synthesis using PSO}
\begin{algorithmic}[1]
    \State \textbf{Input:} $\varphi,\pi_0,\mathbf{S},Stop$, (\textit{hyperparameters} : $W,r_p,r_g,K$)
    \State \textbf{initialize} $[\pi_k,v_k], k = 1,..,K$ \Comment{initialize particle positions and velocities} 
    \State \textbf{while} $\neg Stop$ \Comment{Terminate if stopping condition is met}
        \State \hskip1.5em \textbf{for $k:= 1,..,K$ do}
            \State \hskip1.5em \hskip1.5em  $S_{\pi_k} \gets \mathbf{S}(\pi_k); $  \Comment{Generate trajectories}
            \State \hskip1.5em \hskip1.5em $\pi_k^{best}\gets arg\max_{\pi = \{\pi_k,\pi_k^{best}\}} (\rho (\varphi,S_{\pi})) $
             \State \hskip1.5em \hskip1.5em $ [\pi_k,v_k] \gets $ update according to \eqref{eq:pso1} and \eqref{eq:pso2}
            \State \hskip1.5em $\pi^{best} \gets arg\max_{\pi = \{\pi^{best},\pi_k^{best}|k=1,..,K\}} (\rho (\varphi,S_{\pi}))  $ 
    \State  \textbf{return }$ \pi^* \gets \pi^{best}  $ 
    \end{algorithmic}
\label{alg:par_synth}
\end{algorithm}


\section{Experiments }
\label{sec:experiments}

In this section, we showcase our proposed framework for (1) learning ST specifications for a reaction diffusion system and (2) synthesis of parameters for the same system to produce a desired behaviour. The algorithms are implemented on a PC with a Core i7 CPU @350GHz. For clustering, we used built-in MATLAB functions (\textit{kmeans}). For optimization, we used a custom Particle Swarm Optimization (PSO). The decision tree-based SVM-STL learning algorithm are implemented in Python 3 on an Ubuntu 18.04 system with an Core i7 @3.7GHz and 16GB RAM.

Similar to~\cite{spatel}, we consider a $32 \times 32$ reaction-diffusion system $\mathbf{S}_{RD}$ ~\cite{turing1990chemical} with two species. The concentrations of the species evolve according to $\frac{dx^1_{i,j}}{dt}  = D_1 (\mu^1_{i,j} - x^1_{i,j})+ R_1 x^1_{i,j}x^2_{j,i} - x^1_{i,j} +R_2$
and $\frac{dx^2_{i,j}}{dt}  = D_2 (\mu^2_{i,j} - x^2_{i,j})+ R_3 x^1_{i,j}x^1_{j,i} - x^2_{i,j} +R_4$,
where $x^1_{i,j}$ and $x^2_{i,j}$ are the concentrations of the two species at location $(i,j)$, $\mu^1_{i,j}$ and $\mu^2_{i,j}$ are the inputs to location $(i,j)$ from neighboring locations, i.e. $ \mu^n_{i,j} = \frac{1}{|\upsilon_{i,j}|} \sum_{\upsilon \in \upsilon_{i,j}} x^n_{v}$ with $\upsilon_{i,j}$ being the set of adjacent location indices to location $(i,j)$, $D_1,D_2$ are the diffusion coefficients, and $R_1 =1 ,R_2 = -12,R_3=-1 ,R_4=16 $ are the parameters defining local dynamics for the species. The training set was generated using diffusion coefficients $D_1,D_2$ from the set $P_1 \times P_2$ where $P_1,P_2 = \{0.1,0.2,\ldots, 9.9\}$. 

\textbf{Learning ST properties from system executions: } We consider a set $\mathbf{S} = \{S^{(i)}\}^{N_S}_{i=1}$, where $S^{(i)}: \mathbb{T} \rightarrow \mathbb{R}^{32 \times 32}$ are the ST trajectories, with $\mathbb{T} = [0,60]$ and $N_S= 25000$.

\begin{figure}[htb]
\begin{center}
\includegraphics[width=.99\linewidth]{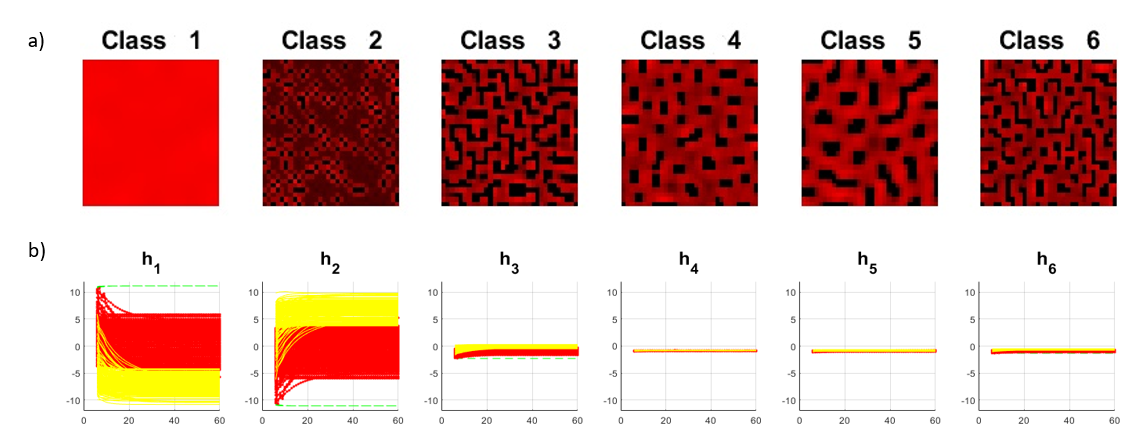}
\caption{\footnotesize a) Sample images from spatial classes $1,..,6$; b) ST trajectories from classes $1,2,3$, color coded by green, yellow and red, respectively.}
\label{fig:clustering}
\end{center}
\end{figure}

To cluster the images in the set $\mathbf{S}$, we use the approach detailed in Sec.~\ref{sec:cluster_images}. We start by removing temporal dependency and create a new set $I = \{I^{(i)}\}_{i=1}^{N_I}$, where $N_I = N_S \times T = 25000 \times 60 = 1500000$. We consider the feature extractor $f_{cnn}$ based on the VGG16 architecture pre-trained on the ImageNet dataset and the distance measure $d_I$. The number of classes is tuned empirically as $n = 6$. Sample images from the spatial classes are shown in Fig.~\ref{fig:clustering}a. Next, we follow the approach presented in Sec.~\ref{sec:learn_predicate} to learn the predicate functions $h_1,..,h_6$ from the labeled set of images $\mathbf{\mathcal{I}}$. A graphical representation is shown in  Fig.~\ref{fig:clustering}b.

\vspace{-5mm}
Using the set $\mathbf{S} = \{S^{(i)}\}^{N_S}_{i=1}$, we cluster the ST trajectories into $n_S$ clusters. We follow the approach detailed in Sec.~\ref{sec:cluster_ST} to cluster ST trajectories into $n_S = 3$ classes. The parameter $n_S$ is tuned manually, by visualising trajectories within each cluster for different values of $n_S$. The outcome of the clustering process is the labeled dataset $\mathcal{S} = \{(s^{(i)},l_S^{(i)})\}_{i=1}^{N_S}$ that is used to learn the SVM-STL specifications.

We solve 3 different two-class classification problems, based on the approach in Sec.~\ref{sec:learn_SVM_STL}, to learn an SVM-STL formula for each class in $\mathcal{S}$.
The hyperparameters and performance metrics are summarized in Tab.~\ref{tab:learning_SVM_STL}. As an example formula, the learned specification for the ST class 3 in one of the folds is $\varphi_3 = \varphi_{31}^{2.8} \wedge \varphi_{32}^{0.6} \wedge \varphi_{33}^{0.1}$, where $\varphi_{31},\varphi_{32},\varphi_{33}$ correspond to decision trees $1,2,3$ and the superscript corresponds to the weight (importance) of the trees. Formulae  $\varphi_{31},\varphi_{32},\varphi_{33}$ are given by: 
\begin{align*}
    \varphi_{31} &= (G_{[19, 49]} h_1 \leq -4.0 \wedge G_{[36, 47]} h_2 > 3.1) \vee (\neg G_{[19, 49]} h_1 \leq -4.0 \wedge F_{[38, 40]} h_2 > 4.2 ) \\ 
    \varphi_{32} &= (G_{[17, 48]} h_1 \leq -4.0 \wedge F_{[22, 44]} h_2 > 3.1) \vee (\neg G_{[17, 48]} h_1 \leq -4.0 \wedge F_{[22, 48]} h_2 > 4.2) \\     
    \varphi_{33} &= (G_{[33, 49]} h_1 \leq -4.1 \wedge G_{[7, 44]} h_2 > 2.8 ) \vee (\neg G_{[33, 49]} h_1 \leq -4.1 \wedge G_{[31, 44]} h_2 > 4.1)
\end{align*}
\vspace{-10mm}

\begin{table}[htb!]
\centering
\label{tab:learning_SVM_STL}
\caption{\footnotesize  Performance summary. Tree depth = 2, Num. of decision trees = 3, K-fold = 2}
\begin{tabular}{|c|c|c|c|}
\hline
Specification &  \begin{tabular}[c]{@{}c@{}}Avg. accuracy (\%)\\ training/testing \end{tabular} & \begin{tabular}[c]{@{}c@{}}Std. dev. (\%) \\ training/testing \end{tabular} & \begin{tabular}[c]{@{}c@{}}Run time\\ (hrs)\end{tabular} \\ \hline
$\varphi_1$      & 99.99 / 99.68  & 0.00/0.00   & 2.1 \\ \hline
$\varphi_2$      &  93.84 / 88.11  & 6.82/15.8    & 1.8 \\ \hline
$\varphi_3$      & 99.48 / 99.24  &  0.26/0.28   & 2.0 \\ \hline
\end{tabular}
\end{table}

The learned formulae are intuitive and easy to interpret. For example, consider $G_{[19, 49]} h_1 \leq -4.0 \wedge G_{[36, 47]} h_2 > 3.1$ from $\varphi_{31}$ above. This specification can be translated to human language as "always in time interval [19,49] the spatial class 1 is not observed AND in time interval spatial class 2 is observed"(see Fig.~\ref{fig:clustering}a). The thresholds for predicates $h_1 \leq -4.0$ and $h_2 > 3.1$ show how strong the spatial classes $1,2$ are met (see red lines in Fig.~\ref{fig:clustering}b)

\textbf{Parameter Synthesis:} we applied the parameter synthesis approach presented in Alg.~\ref{alg:par_synth} to find a pair of diffusion coefficients that maximize the degree of satisfaction, with respect to the requirements given by the SVM-STL formula: 
$\psi = F_{[0,30)}G_{[0,60)} h_5(S)>0 \wedge G_{[0,60)} h_4(S)< 0$

\begin{figure}[htb]
\begin{center}
\includegraphics[width=.9\linewidth]{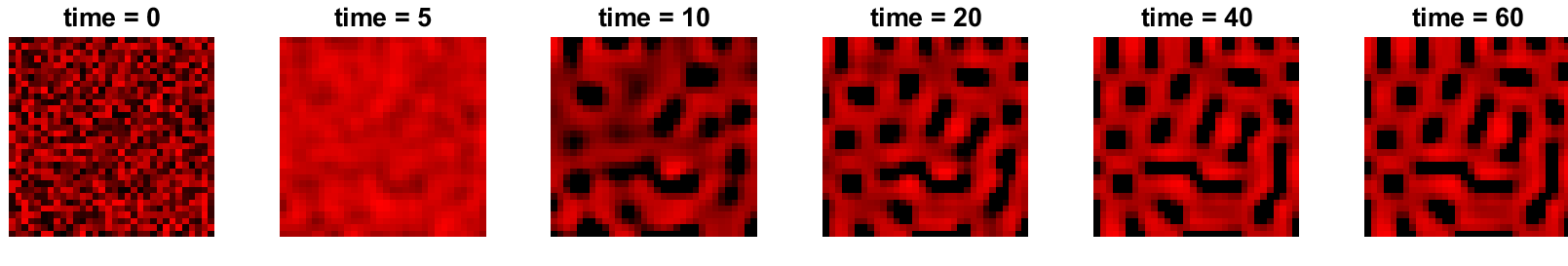}
\caption{\footnotesize Sample trajectory $S \models \psi= F_{[0,30)}G_{[0,60)} h_5(S)>0 \wedge G_{[0,60)}h_4(S)< 0$, where $h_5$ corresponds to class 5 \textit{{large spots}} and $h_4$ corresponds to Class 4 \textit{{small spots}}, using parameters $D_1 = 3.9$ and $D_2= 30 $}
\label{fig:par_synth}
\end{center}
\end{figure}

Using Alg.~\ref{alg:par_synth} with with $K = 100, W = 0.6,r_p = 1.5,r_g = 2.5$ and a max number of iterations $20$; we found parameters $D_1 = 3.9 , D_2 = 30$ that result in satisfying the formula $\psi $. PSO found a ST trajectory $S$ (see Fig.~\ref{fig:par_synth}) the satisfies $\psi$ in $\sim 2.5$ minutes with a robustness score $\rho(S,\psi,0) = 0.11$. The results illustrate the capability of SVM-STL to specify a wide range of spatial and temporal requirements for dynamical systems, and synthesizing parameters to meet them.

\vspace{-12mm}
\section{Conclusions and Future Work}

In this work, we investigated the problem of learning spatial temporal logic formulas from time-trajectories of images. We introduced SVM-STL, an extension of Signal Temporal Logic that allows for specifying spatial properties. Our framework can learn SVM-STL formulas from labeled as well as unlabeled data. We also presented a method to synthesize the parameters for a dynamical system from such specifications. The learning framework and the parameter synthesis method were showcased on a reaction-diffusion system. In future research, we will explore the idea of learning specifications using end-to-end CNN models as predicates. Such models would take one (or multiple) images as inputs and would output correct robustness scores for each spatial class. We will also investigate alternative approaches for clustering high dimensional spatio-temporal signals. 


\balance


\bibliography{main}

\end{document}